# A Graph-Based Inference Method for Conditional Independence

Ross D. Shachter
Department of Engineering-Economic Systems
Stanford University
Stanford, CA 94305-4025
shachter@sumex-aim.stanford.edu

## Abstract

The graphoid axioms for conditional independence, originally described by Dawid [1979], are fundamental to probabilistic reasoning [Pearl, 1988]. Such axioms provide a mechanism for manipulating conditional independence assertions without resorting to their numerical definition. This paper explores a representation for independence statements using multiple undirected graphs and some simple graphical transformations. The independence statements derivable in this system are equivalent to those obtainable by the graphoid axioms. Therefore, this is a purely graphical proof technique for conditional independence.

## 1. INTRODUCTION

The graphoid Axioms for conditional independence were originally proposed as the fundamental theme behind statistical inference[Dawid, 1979]. Since then, they have come to be accepted as a complete characterization and definition of conditional independence, applicable to many situations far beyond probabilistic models [Fagin, 1977; Geiger and Pearl, 1990b; Pearl, 1988; Pearl et al., 1990; Smith, 1989; Smith, 1990; Verma and Pearl, 1988].

This article explores a graphical way of representing independence statements using multiple undirected graphs first suggested in [Shachter, 1990]. Its main contribution is showing that repeatedly applying two graphical transformations, node deletion and graph combination, is equivalent to repeatedly applying the graphoid axioms. Consequently a purely graphical method of deriving new independence statements from a given set of such statements is obtained. The graphical operations have the advantage of an intuitive representation in undirected graphs, which makes the structure of independence assumed in a model explicit. They have the potential for making the abstract notion of conditional independence more accessible for teaching and knowledge acquisition. Nonetheless, although the technique is simple to use manually, its complexity is exponential, as are the derivation steps using the graphoid axioms.

The key to these graphical operations is the multiple undirected graph framework developed by Paz[1987; 1988] and Geiger[1987]. Many people have looked at undirected graph representations for the graphoid axioms, but they have always kept the intersection axiom, restricting the applicability to purely positive probability distributions and limiting the possible generalizations beyond probability [Pearl, 1988]. The key result underlying this paper is the development of a system which incorporates the contraction axiom but does not incorporate the intersection axiom.

Section 2 presents the notation and basic principles of the graphoid axioms while Section 3 defines the framework of Multiple Undirected Graphs and the graphical transformations on them. Section 4 proves the equivalence between the two axiom systems. Section 5 extends the graphical operations to some important special cases while Section 6 presents examples of theoretical properties, practical applications, and efficient computational structures which follow from the axioms.

## 2. NOTATION AND BASIC CONCEPTS

The notation and framework used throughout the paper is described in this section.

The primitive entities in a model are a finite set of **elements** U. In a probabilistic model, these correspond to random variables. A single element or set of elements will be denoted by a capital letter, such as X, Y, or Z. These elements are associated with nodes in an undirected graph. For Sections 3 and 4, it will be assumed that each node corresponds to a single element, but in Section 5 this will be generalized to allow multiple elements to be associated with a single node. For readability, a node will be referred to by the element(s) associated with it. For example, if element X is associated with a particular node, then that node will also be called X.

A **dependency model** over a finite set of elements U is a three place predicate I( X, Z, Y ) where X, Z, and Y are disjoint subsets of U. The intended interpretation of I( X, Z, Y ) is that X is conditionally independent of Y given Z. Alternatively, having observed Z, no additional information about X could be obtained by also observing Y. For example, in a probabilistic model, I( X, Z, Y ) holds if and only if

$$P\{ X \mid Z, Y \} = P\{ X \mid Z \} \text{ whenever } P\{ Z \} > 0$$

for every value of the variables X, Y, and Z.

The realization of I need not necessarily be probabilistic [Fagin, 1977; Geiger and Pearl, 1990a; Geiger and Pearl, 1990b; Geiger et al., 1990; Pearl, 1988; Pearl et al., 1990; Smith, 1989; Smith, 1990; Verma and Pearl, 1988]. All of the those formulations satisfy the following **graphoid** axioms, first proposed by Dawid[1979]:

**Symmetry:** I( X, Z, Y ) ⇔ I( Y, Z, X ) ;

**Decomposition:** I( X, Z, Y ∪ W ) ⇒ I( X, Z, Y ) ;

**Weak Union:** I( X, Z, Y ∪ W ) ⇒ I( X, Z ∪ Y, W ) ;

**Contraction:** I( X, Z ∪ Y, W ) and I( X, Z, Y )
$$\Rightarrow I( X, Z, Y \cup W ).$$

The essence of these axioms is that learning an irrelevant proposition does not change the status of other facts; every proposition that was irrelevant remains irrelevant and every proposition that was relevant remains relevant. These axioms are not complete for probabilistic independence [Geiger and Pearl, 1990a], but are nevertheless powerful enough to derive useful consequences that generalize from probabilistic independence. We shall next used sets of undirected graphs for deriving such consequences.

## 3. MULTIPLE UNDIRECTED GRAPHS

The representation of independence by Multiple Undirected Graphs was studied by Paz[1987; 1988] and Geiger[1987]. We will see below how it can be used as a visual technique for deriving new independence statements from a given list of such statements.

Conditional independence is represented in an undirected graph by graph separation: given three disjoint sets of elements from U, X, Y, and Z, Z **separates** X from Y if every path between X and Y contains an element from Z. For example, Z separates X from Y in the graph shown in Figure 1a representing I( X, Z, Y), but there is no independence represented in the graph shown in Figure 1b.

In this section and the following section, it is assumed that each node corresponds to exactly one element. In section 5 this will be generalized to allow multiple elements in a node, and the graph separation rule will apply even when the sets X, Y, and Z are not disjoint.

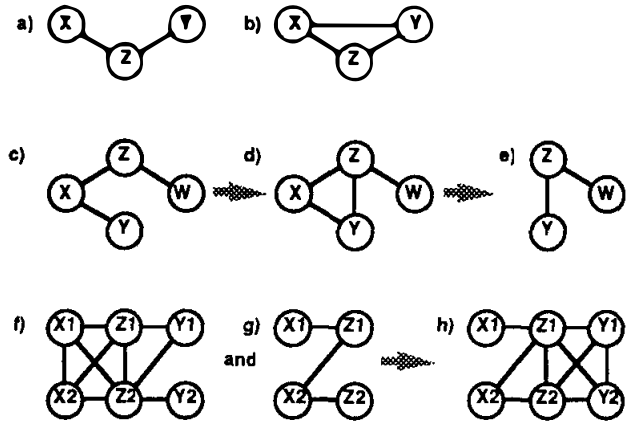

Figure 1: MUGs and Their Basic Operations

A **multiple undirected graph** (MUG) M over U is a dependency model consisting of a set of undirected graphs containing elements from U. An independence statement I( X, Z, Y ) holds in M if and only if there exists a graph in M containing X ∪ Y ∪ Z in which Z separates X from Y. Alternatively, I( X, Z, Y ) is said to be **satisfied** by M, or M **satisfies** it. Suppose for example that the graphs shown in Figures 1a and 1b constitute a MUG. Then I( X, Z, Y ) even though there is no independence represented in the second graph.

Two dependency models are **equivalent** if they represent the same list of independence statements. The following transformations add undirected graphs to a given MUG, producing a new MUG that is equivalent to the original one. Given any graph G in a MUG, then the MUG obtained by duplicating G and adding any arcs is said to be derived by **arc addition**. Given any graph G in a MUG M, then the MUG obtained from M by duplicating G and deleting any node after adding arcs between all of the nodes which had been adjacent to it is said to be derived by **node deletion**.

Both arc addition and node deletion follow directly from graph separation, since any separation present in the new graph must also have been present in the original graph. In the example graph drawn in Figure 1c an arc is added between Z and Y to obtain the graph shown in Figure 1d. Node X can be deleted from the graph shown in Figure 1c to create the graph shown in 1e, but only after an arc is added between Z and Y, the nodes adjacent to X, to obtain the graph shown in Figure 1d. Although X separates Y from Z ∪ W, that separation can no longer be represented once X is deleted. Instead the fact that there is a path from Z to Y through X requires that X and Y be connected in the new graph. This leads to the following result.

**Theorem 1.** Let M be a MUG. Then a MUG derived from M by node deletion or arc addition is equivalent to M.

Next we define a new transformation which does not produce an equivalent MUG because it adds independence statements not represented in the original MUG. Let M be a MUG over U, and X, Y, and Z be disjoint subsets of U. If M satisfies I( X, Z, Y ) and there is a graph G in M for which X ∪ Z is the set of all of the nodes, then the MUG obtained from M by duplicating G, adding to the new graph nodes for each element in Y, and connecting every pair of nodes in Y ∪ Z with an arc is said to be derived by **graph combination**.

Graph combination synthesizes a new graph from two others. Consider the graph drawn in Figure 1f in which (Z1 ∪ Z2) separates (X1 ∪ X2) from (Y1 ∪ Y2), while the graph drawn in Figure 1g contains only the elements (X1 ∪ X2) ∪ (Z1 ∪ Z2). Using graph combination, a new graph is formed as shown in Figure 1h. In the process, arcs are added between all of the nodes in (Y1 ∪ Y2) ∪ (Z1 ∪ Z2). Note that Z1 separates X1 from (Y1 ∪ Y2) which could not be determined from either of the original graphs.

## 4. GRAPHICAL REPRESENTATION OF THE GRAPHOID AXIOMS

The thrust of this paper is to show that graph combination and node deletion add all and only those independence statements derivable by the graphoid axioms. Consequently, these transformations provide a graph-based technique for proving all independence statements implied by the graphoid axioms (and none other) from a given input set of independence statements. The input for our proof technique can be specified either by a list of undirected graphs, or a list of independence statements, or a combination thereof. This proof technique is quite useful because the graphoid axioms are strong enough to prove powerful results [Verma and Pearl, 1988].

**Theorem 2.** Let M be a MUG and let *I* be the set of independence statements satisfied by M. An independence statement I is derivable by the graphoid axioms from *I* if and only if there exists a sequence of node deletions and graph combinations on M that produces a MUG in which I is satisfied.

**Proof:**

Let I be a statement derived from *I* using the graphoid axioms. Let $\sigma_1 ... \sigma_k$ be a derivation chain of I. That is, $\sigma_k$ equals I and each $\sigma_i$ is either an independence statement in *I* or it has been derived by the graphoid axioms from $\sigma_1$ ...$\sigma_{i-1}$. Next, we construct a sequence of graphical transformations that would end with a MUG that satisfies I. The proof is done by induction on k.

When k is one, then since $\sigma_1$ is in *I*, there exists a graph in M where $\sigma_1$ is satisfied. Otherwise, $\sigma_i$ is derived from previous statements by one of the graphoid axioms. Suppose $\sigma_i$ is derived from some $\sigma_j$ (i > j by symmetry. Consider the graph in which $\sigma_j$ is satisfied. The same graph satisfies $\sigma_i$ as well since if Z separates X from V it must also separate V from X.

Otherwise, $\sigma_i$ must have been derived from previous statements using decomposition, weak union, or contraction. To verify that the same statement could be obtained graphically, consider the graphs drawn in Figure 2, representing the graphoid axioms. If the previous statement is I(X, Z, Y ∪ W), as shown in the graph drawn in Figure 2c, then weak union and decomposition , shown in the graphs drawn in Figure 2a and 2b, respectively, are recognized graphically since Z separates X from Y, and Z ∪ Y separates X from W. If the previous statements are I(X, Z ∪ Y, W) and I(X, Z, Y) then a result equivalent to the contraction axiom can be obtained using graph combination. Because I(X, Z ∪ Y, W) and there is a graph containing only X ∪ (Z ∪ Y), that graph can be duplicated, adding to it nodes for W and connecting every pair of nodes in W ∪ (Z ∪ Y) to obtain the graph shown in Figure 2c in which Z separates X from Y ∪ W.

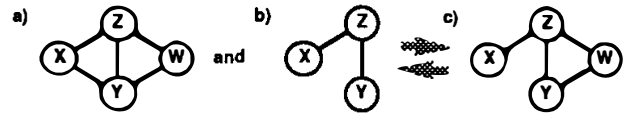

Figure 2: The Graphoid Axioms

Now suppose $M_1 ... M_k$ is a sequence of MUGs, $M_k$ satisfies I, and every $M_i$ is obtained from the previous ones by either node deletion or graph combination. It must be shown that I can be derived by the graphoid axioms from the independence statements encoded in $M_1$, i.e., from *I*. The proof is done by induction on k. The basis k=1 holds since I is satisfied by the given MUG.

Suppose $M_i$ is derived from $M_j$ by node deletion performed on graph G. Let G' be the added graph with a deleted node. Since any statement I represented by G' must also hold in G, it need not be derived.

Similarly, in the case of graph combination, it can be shown that every statement added is derivable using the graphoid axioms. Q. E. D.

## 5. EXTENSIONS TO THE GRAPHICAL OPERATIONS

In this section, the Graphical Axioms are generalized by relaxing the assumptions, made in Section 3, that each node is associated with a single element, and that independence relations are defined only on disjoint sets. These extensions require no changes to the graphical operations, but they greatly increase their power.

First, it is necessary to generalize the graphoid axioms to define independence relations even when the three sets of elements are not disjoint. This is accomplished through an additional axiom [Pearl, 1988]:

**Overlap:** $I(X, Z, Y) \Leftrightarrow I(XZ, Z, YZ)$.

Thus $I(X, Z, Y)$ is possible when X, Y, and Z are not disjoint if $X \cap Y \subseteq Z$. No modifications to the graphical operations are required, since Z separates X from Y if every path between X and Y contains an element of Z. This condition can be satisfied when some of the elements of X and Y are also elements of Z. For example, consider the graph shown in Figure 3a). Since Z separates X from Y it follows that $I(X, Z, Y)$, but also $I(X \cup Z, Z, Y \cup Z)$, $I(X \cup Z, Z, Y)$ and $I(X, Z, Y \cup Z)$. It is not usually possible, however, to separate two overlapping sets of elements unless their intersection is included in the separating set.

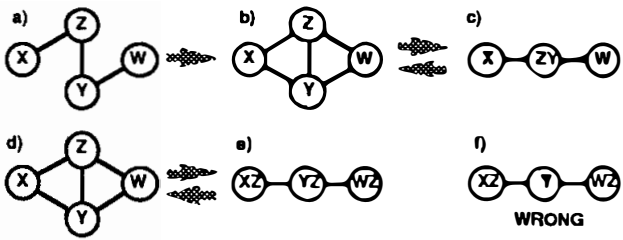

Figure 3: Properties with Multiple Elements in a Node

Now that the definition of independence has been extended, it is possible to have multiple elements be assigned to the same node and also to have the same element assigned to multiple nodes. This leads to two additional graphical transformations. Given any graph G in a MUG M, then the MUG obtained from M by duplicating G and replacing two nodes Z1 and Z2 with one node Z containing the union of the elements associated with Z1 and Z2 is said to be derived by **node merging**. In the new graph, arcs must be drawn between node Z and any node which had been adjacent to node Z1 or node Z2. Given any graph G in a MUG M, then the MUG obtained from M by duplicating G and replacing one node Z with two nodes Z1 and Z2 which together contain all of the elements associated with Z is said to be derived by **node splitting**. In the new graph, arcs must be drawn between nodes Z1 and Z2, and between them and any node which had been adjacent to node Z.

The node merging and node splitting properties follow directly from the extended notion of graph separation, in that any separation present after the change must have been present beforehand. In the case of node merging, separation by the new node is equivalent to separation by both old nodes, while after node splitting, separation by both new nodes is equivalent to separation by the old one. In either case, any path present beforehand will be maintained. Consider the example graphs shown in Figure 3. Arcs are added to the graph drawn in part a) to obtain the graph drawn in part b). The Z and Y nodes are then merged to obtain the graph drawn in part c). The transition from part a) to part c could, of course, be done as a single step. If the ZY node in the graph drawn in part c) is now split, it results in the graph drawn in part b). It would not be possible to infer the graph drawn in part a) from the graph drawn in part c), so conditional independence information is lost in the process of Node Merging. This leads to the following result.

**Theorem 3.** Let M be a MUG. Then a MUG derived from M by node splitting or node merging is equivalent to M.

An example of a graph with repeated elements is shown in the Figure 3e. Note that if the same element appears in two different nodes, then it should appear in every node on a path between them, since it can only be separated from itself by itself, and thus the graph shown in Figure 3f would usually be in error. A graph G with multiple elements in a node can always be transformed into another graph G' with single elements in a node. An arc should be drawn between two nodes in G' if and only if their elements were in the same or adjacent nodes in G.

## 6. EXAMPLES

The graphical operations are useful because they allow us to conceptualize abstract conditional independence and to derive easily results which arise from the graphoid Axioms. These sections describe some different types of results which can be derived using the graphical operations.

### 6.1 ADDITIONAL PROPERTIES

Pearl [Pearl, 1988] describes some additional properties related to conditional independence. The graphical operations allow us to derive easily those which apply and to recognize those which do not. These properties, in independence notation are:

**Mixing:** I( X ∪ Y, Z, W ) and I( X, Z, Y )
⇔ I( X, Z, Y ∪ W );

**Chaining:** I( X ∪ Z, Y, W ) and I( X, Z, Y )
⇔ I( X, Z, W ); and

**Intersection:** I( X, Z ∪ Y, W ) and I( X, Z ∪ W, Y )
⇐ I( X, Z, Y ∪ W ).

We consider each in turn using the graphs drawn in Figure 4.

The Mixing Property assumptions are represented in the graphs drawn in parts a) and b). Using graph combination, the graph drawn in part c) can be constructed from the one in part b) by adding node W and an arc between Z and W. The resulting graph not only shows that Z separates X from Y ∪ W, but also the symmetry by which Z separates all of the elements from each other. It is simple to verify that the graphs shown in parts a) and b) can be obtained from the graph shown in part c) by either arc addition or node deletion. Mixing is a useful property since it together with symmetry and decomposition have been shown to completely characterize marginal independence [Geiger et al., 1991].

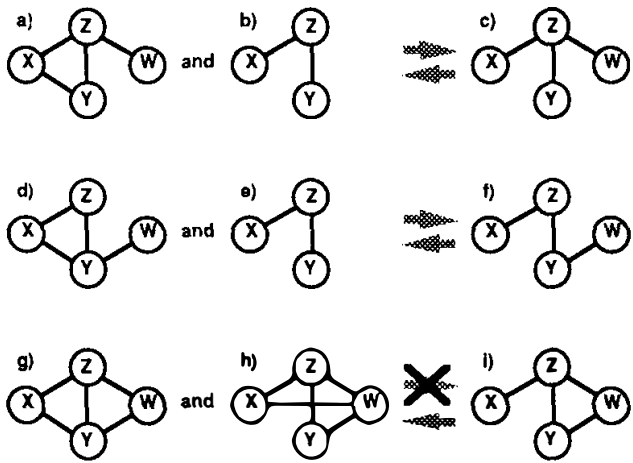

Figure 4: Additional Properties on MUGs

The Chaining Property conditions are represented in the graphs drawn in parts d) and e). Again graph combination allows the graph drawn in part f) to be constructed from the one in part e) by adding node W and an arc between Y and Z. Clearly Z separates X from W in the new valid graph, but it is also more apparent why this is called "chaining." It is again simple to obtain the graphs shown in parts d) and e) from the graph shown in part f) by arc addition and node deletion.

Finally, the conditions for the Intersection Property are satisfied by the graphs drawn in parts g) and h). However, graph combination can not be applied directly, since both graphs are on the same elements, X ∪ Y ∪ Z ∪ W. If any nodes are deleted from either of the graphs, there is no conditional independence remaining, and it is not possible to obtain the graph shown in part i). This is as it should be, because the Intersection Property is not true in general, but rather requires further assumptions. (In the case of probabilistic models, for example, all of the joint probabilities must be strictly positive.) The graph shown in part i) could be obtained if the conditions corresponded to the graphs shown in parts g) and e), rather than g) and h). On the other hand, the graphs shown in parts g) and h) can be obtained from the graph shown in part i) through arc addition.

## 6.2 DIRECTED GRAPHS, MORAL GRAPHS, AND JOIN TREES

In practice, probabilistic models are most easily assessed in the form of a directed graph, called an influence diagram or belief network [Howard and Matheson, 1984; Pearl, 1986b]. For most efficient computation these models can then be converted into undirected graphs, called **moral graphs** [Lauritzen and Spiegelhalter, 1988], and then into trees of overlapping sets of nodes, called **join** trees [Beeri et al., 1983; Jensen et al., 1990a; Jensen et al., 1990b]. In this section, it is assumed that the reader is familiar with these concepts, so that the focus will be on insights to be gained from the Graphical Axioms.

The directed graph can be thought of as a sequence of conditional independence statements: each element is conditionally independent of the elements listed before it, given its parents in the graph. In fact, the conditional independence does not apply to any particular sequence, but rather to any ordering consistent with the graph. This property is well known [Pearl, 1988; Smith, 1989], but it can also be proven directly through the Graphical Axioms [Shachter, 1990].

This sequence of conditional independence statements can be used to construct the moral graph through repeated application of the Combination Property. Each new element can be added in turn to the undirected graph under construction by adding its node and arcs between it and its parents and among its parents. (This is the marrying of the parents which gives "moral" graphs their name.) Thus the directed graph shown in Figure 5a) has the moral graph shown in part b) (ignoring the dashed (L, B) arc to be discussed below). Moralizing arcs (T, L) and (E, B) are added between the parents of E and D, respectively. Note that we really have a sequence of moral graphs, including the one shown in part c), revealing independence not shown in the full moral graph. This is why the MUG representation is needed in general.

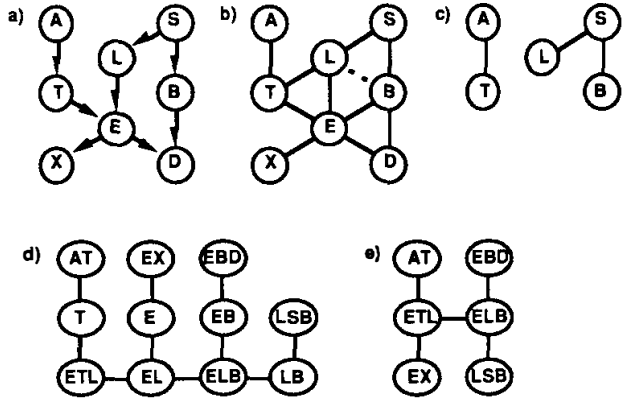

Figure 5: Directed Graph, Moral Graphs, and Join Trees

For efficient inference, it is ideal to have an undirected graph in the form of a tree [Jensen et al., 1990a; Jensen et al., 1990b; Lauritzen and Spiegelhalter, 1988; Pearl, 1988]. In general this cannot be accomplished with nodes corresponding to single elements, but instead requires nodes with sets of elements and some elements assigned to multiple nodes. Such trees are shown in parts d) and e). They are join trees, because whenever an element appears in two different nodes, it appears in every node on the path between them. Recall that this is just the necessary property for assignment of elements to multiple nodes described in Section 5. Note that, given the independence represented in the directed graph, the join tree is a valid graph. To capture all of that independence, however, we need the graph in part d) rather than the one in part e). Notice that through the insertion of *separation sets*, containing the intersection of neighboring node sets, we create a graph in which the only arcs needed for a single element graph such as the one in part b) are between elements in the same node in part d). When we create the graph in part b) in this manner, it has become a *chordal graph* with the addition of the dashed arc (L, B) [Beeri et al., 1983].

### 6.3 DETERMINISTIC ELEMENTS AND SEPARATION

The independence properties in the directed graph have been extensively studied [Geiger et al., 1989; Geiger et al., 1990; Pearl, 1986a; Pearl, 1988; Shachter, 1988; Shachter, 1990; Verma and Pearl, 1988], but recent work has shown how they can be recognized in the undirected graph [Lauritzen et al., 1990]. Unfortunately, one type of independence appears difficult to represent in the undirected graph: elements which are deterministically related to other elements. Nonetheless the results of Lauritzen et al[1990] can be extended in the undirected graph to recognize independence in models with deterministic elements.

An element in a directed graph is said to be <u>deterministic</u> and drawn with a double oval if it is conditionally independent of all elements, including itself, given its parents. In a probabilistic model, such an element can be described as a deterministic function of its parents. For example, in the diagram shown in Figure 6a), the element B is a deterministic function of A. When an element is deterministic, an operation called <u>deterministic propagation</u> [Shachter, 1988; Shachter, 1990] can be applied to the graph: the children of the deterministic node have their arc from the deterministic node replaced by arcs from its parents. For example, after deterministic propagation, the graph drawn in part a) is transformed into the graph drawn in part b). Such an operation can be interpreted in probabilistic models as substitution of the deterministic function into the children's distributions, but it can also be derived in general using the graphical operations [Shachter, 1990]. In the case of the example, we can justify deterministic propagation by considering the moral graphs for the graphs in parts a) and b) shown in parts c) and d), respectively. To obtain the moral graph shown in part d) from the one shown in part c), first delete the node B, and then use the Combination Property, recognizing that because node B is deterministic, I( B, A, C ∪ D ).

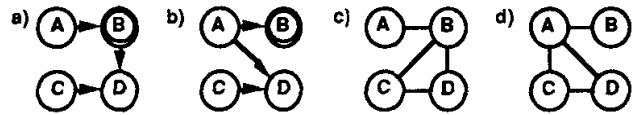

Figure 6: Managing Deterministic Elements

Using the property of deterministic propagation, the independence test of Lauritzen et al[1990] can be generalized to problems with deterministic nodes, to test for what is called D-separation [Geiger et al., 1990; Shachter, 1988; Shachter, 1990]. The procedure to test whether I( X, Z, Y ) is satisfied in a directed graph, possibly containing deterministic nodes, is as follows:

1. Discard elements in the directed graph which are not in X ∪ Y ∪ Z or one of their ancestors in the directed graph.

2. Visiting each element in graph order, perform deterministic propagation on any deterministic element which is not in Z.

3. Form the moral graph of the resulting directed graph.

4. Determine whether Z separates X from Y in the moral graph.

This procedure is applied to some examples as shown in Figure 7. First consider the directed graph shown in part a). There are no deterministic elements and the corresponding moral graph is shown in part b). Clearly I( W, Z, X ∪ Y ∪ V ). The entire moral graph is needed to test the independence of element V from other elements, and Z does not separate X from Y ∪ V. However, to just check whether I( X, Z, Y ), the element V is discarded from the graph shown in part a) and the moral graph is now the one shown in part c). In fact, in this case, I( X, Z, Y ). Another example is the directed graph shown in part d). Its moral graph is shown in part e) and clearly it does not satisfy I( X, Z, Y ). If Z were not observed the we would obtain the moral graph shown in part f), so I( X, ∅, Y ) is true.

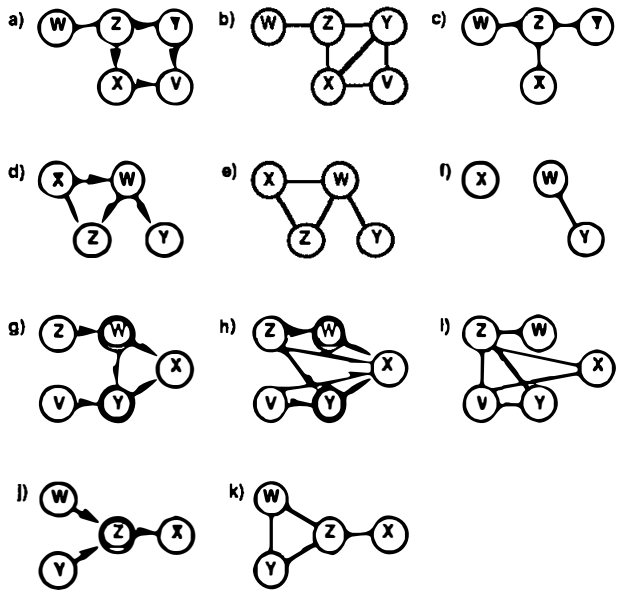

Figure 7: Examples of D-Separation

The next two examples involve deterministic elements. In the directed graph shown in part g), there are two deterministic elements, neither of which is an element in Z. Deterministic propagation results in the directed graph shown in part h), in which arc (W, Y) has been replaced by (Z, Y), (W, X) has been replaced by (Z, X), and (Y, X) has been replaced by (V, X) and (Z, X). The moral graph corresponding to this directed graph is shown in part i). Note that while Z separates W from X ∪ Y ∪ V, Z does not separate X from Y. This is due to the uncertainty introduced into element Y from element V. Finally, although there is a deterministic element in the directed graph shown in part j), deterministic propagation should not be performed because the element is contained in Z.

In the corresponding moral graph shown in part k), Z separates X from Y ∪ W. If deterministic propagation had been performed, this independence would not have been recognized.

## 7.    CONCLUSIONS

In this paper, system of graphical transformations on Multiple Undirected Graphs are defined and shown to be equivalent to the graphoid axioms. These axioms have become accepted as the fundamental properties of conditional independence generalized from probabilistic models. These graphical operations are illustrated with examples related to theoretical properties, practical applications, and efficient computations. This representation facilitates the communication and development of intuition for the abstract definition of conditional independence. Although stress has been placed in this paper on the relative benefits of the new approach, its equivalence to the graphoid axioms is especially powerful in that it allows one to use whichever system is convenient for the problem at hand. Unfortunately, the complexity of both methods is exponential.

The essence of this paper is that graph based techniques can help us to reason about independence relations. Pearl and others have shown a dual aspect by which independence helps us to characterize graphical representations. Both views are useful for exploring the connections between separation in graphs and independence in probability.

### Acknowledgements

This paper benefitted greatly from the comments of Azaria Paz, Judea Pearl, and an anonymous referee, but it is dedicated to Danny Geiger, without whom it would not exist. He not only encouraged me to pursue this research question in the first place, but he was also most generous with his suggestions and patient tutoring.